\documentclass[conference]{IEEEtran}
\IEEEoverridecommandlockouts
\usepackage{cite}
\usepackage{amsmath,amssymb,amsfonts}
\usepackage{algorithmic}
\usepackage{graphicx}
\usepackage{textcomp}
\usepackage{xcolor}

\usepackage{cuted}
\usepackage{caption}
\usepackage{hyperref}
\usepackage{algorithm}
\usepackage{enumitem}
\usepackage{booktabs}

\usepackage{eqparbox}

\def\BibTeX{{\rm B\kern-.05em{\sc i\kern-.025em b}\kern-.08em
    T\kern-.1667em\lower.7ex\hbox{E}\kern-.125emX}}
\begin{document}

\title{PromptArtisan: Multi-instruction Image Editing in Single Pass with Complete Attention Control}

\author{
	\IEEEauthorblockN{Kunal Swami, Raghu Chittersu, Pranav Adlinge, Rajeev Irny, Shashavali Doodekula, Alok Shukla}
	\IEEEauthorblockA{
		\textit{Samsung Research India Bangalore}\\
	}
	\{kunal.swami, raghu.c, p.adlinge, rajeev.i, shasha.d, alok.shukla\}@samsung.com
}

\maketitle

\begin{strip}
	\begin{minipage}{\textwidth}
		\centering
		\includegraphics[width=0.86\textwidth]{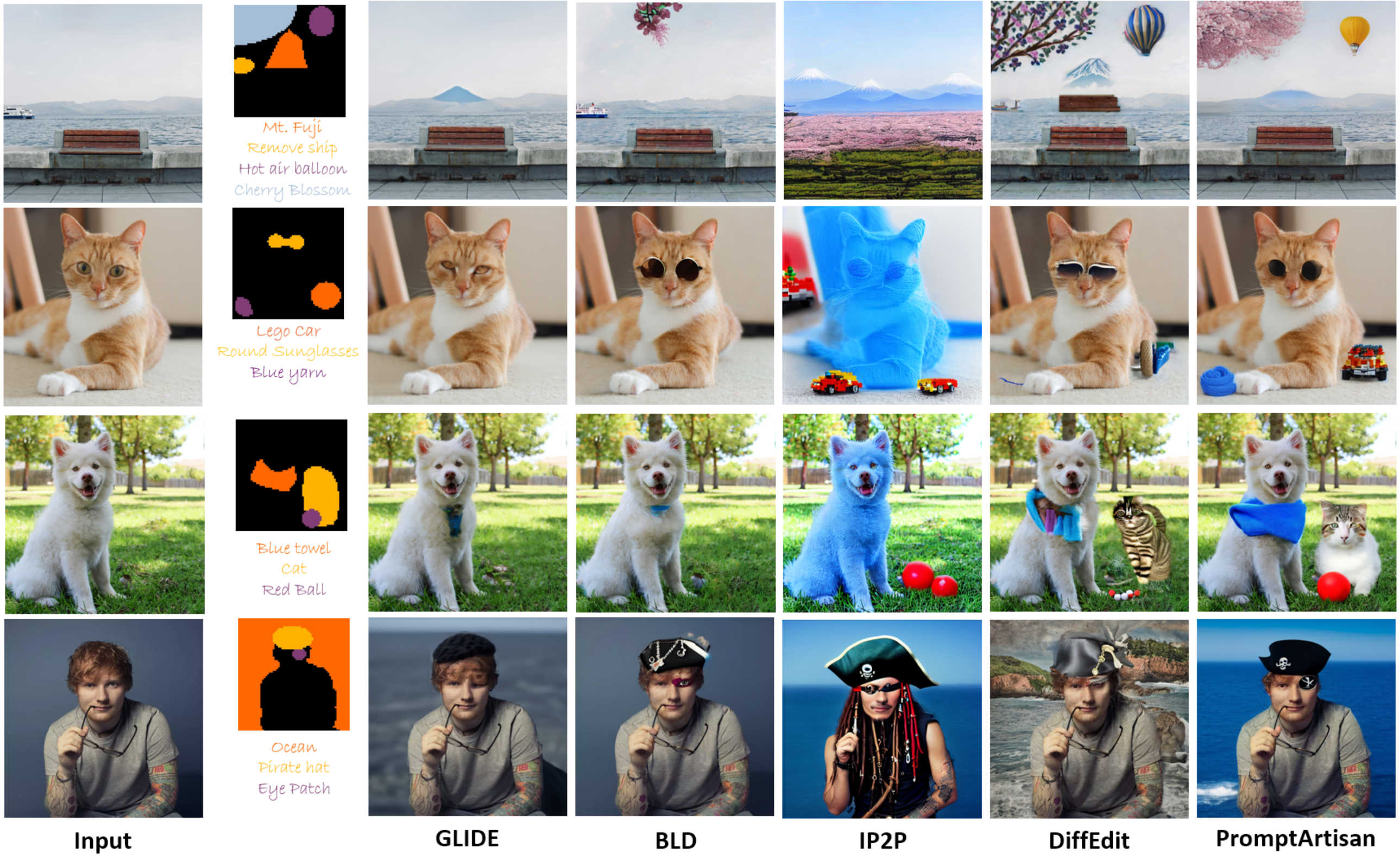}
		\vspace{-4pt}
		\captionof{figure}{PromptArtisan enables single pass editing with multiple mask-prompt pairs. Additionally, multiple mask-prompt pairs enable unprecedented flexibility and complex edits with intersections or overlaps of masks.}
		\label{fig:teaser}
		\vspace{-12pt}
	\end{minipage}
\end{strip}

\begin{abstract}
We present PromptArtisan, a groundbreaking approach to multi-instruction image editing that achieves remarkable results in a single pass, eliminating the need for time-consuming iterative refinement. Our method empowers users to provide multiple editing instructions, each associated with a specific mask within the image. This flexibility allows for complex edits involving mask intersections or overlaps, enabling the realization of intricate and nuanced image transformations. PromptArtisan leverages a pre-trained InstructPix2Pix model in conjunction with a novel Complete Attention Control Mechanism (CACM). This mechanism ensures precise adherence to user instructions, granting fine-grained control over the editing process. Furthermore, our approach is zero-shot, requiring no additional training, and boasts improved processing complexity compared to traditional iterative methods. By seamlessly integrating multi-instruction capabilities, single-pass efficiency, and complete attention control, PromptArtisan unlocks new possibilities for creative and efficient image editing workflows, catering to both novice and expert users alike.
\end{abstract}

\begin{IEEEkeywords}
Multi-instruction Editing, Single Pass Editing, Diffusion Models, Generative AI
\end{IEEEkeywords}

\section{Introduction}
\label{sec:intro}
Manual image editing often demands substantial time investment and proficiency in specialized software tools. Generative AI, particularly with the advancements in diffusion models \cite{ddpm_ho_neurips2020,diffusion_beats_gans_neurips2021,stablediffusion_cvpr2022,imagen_neurips2022,sdxl_arxiv2023,stablecascade_iclr2024}, offers the potential to democratize this capability. Instruction-based image editing (IBE) models \cite{ip2p_cvpr2023,magicbrush_neurips2023,mdp_arxiv2023,zero_shot_i2i_siggraph2023,ibe_mllm_arxiv2023,smartedit_cvpr2024,instructdiffusion_arxiv2023,masactrl_iccv2023,plugnplay_diffusion_cvpr2023}, built upon these generative models, enable users to execute various edits through textual prompts, such as style transfer, subject modifications, or adjustments to colors and expressions.

Despite these impressive strides, existing IBE methods exhibit limitations that motivate our work:

\begin{enumerate}[leftmargin=*]
	\itemsep0em
	\item \textbf{Text-based editing:} Current IBE techniques rely solely on textual instructions \cite{ip2p_cvpr2023,magicbrush_neurips2023,mdp_arxiv2023,zero_shot_i2i_siggraph2023,ibe_mllm_arxiv2023,smartedit_cvpr2024,instructdiffusion_arxiv2023}, hindering precise localization of edits within an image. Furthermore, text-based descriptions can prove ambiguous in scenarios involving multiple instances of a subject or complex editing requirements.
	\item \textbf{Sequential editing:} Existing methods typically constrain users to perform one edit at a time, necessitating sequential execution for multiple modifications. This inherently introduces inefficiencies and potential inconsistencies in the final edited image.
\end{enumerate}

While mask-based image editing techniques \cite{glide_icml2022,blended_diffusion_cvpr2022,smartbrush_cvpr2023,blended_latent_diffusion_siggraph2023} offer precision in attribute modification,
they are limited to single edits per execution. Furthermore, these methods either necessitate retraining for additional mask inputs \cite{glide_icml2022,smartbrush_cvpr2023} or rely on computationally expensive test-time optimization for zero-shot capabilities \cite{blended_diffusion_cvpr2022,blended_latent_diffusion_siggraph2023}.

Consequently, existing IBE methods fall short in addressing the needs of users who require simultaneous and precise editing of multiple attributes. The current alternative, sequential editing, incurs a substantial computational burden due to iterative processing through the diffusion model. Additionally, repeated auto-encoding \cite{vae_iclr2014,cad_vae_arxiv2023} or artifact generation during successive edits can compromise fine-grained image details and overall aesthetics.

To overcome these limitations, we propose a novel and pragmatic problem setting: \emph{enabling simultaneous, multi-attribute image editing with precise spatial control}. Our framework utilizes multiple mask-prompt pairs, affording users unprecedented flexibility and facilitating complex edits involving mask intersections or overlaps---capabilities beyond the reach of current methods. To achieve this, we introduce a \emph{Complete Attention Control Mechanism (CACM)} that modulates both cross-attention (pixel-prompt interaction) and self-attention (pixel-pixel interaction) within the diffusion model. Crucially, our approach operates in a zero-shot manner, eliminating the need for any additional training or test-time optimization of the base generative model.

Given the novelty of our problem setting, we introduce a new benchmark dataset, MiE-Bench, designed to rigorously evaluate and compare methods in this domain.

Following are the \textbf{major contributions} of our work:

\begin{itemize}[leftmargin=*]
	\item We introduce a novel problem setting that enables simultaneous, multi-instruction image editing with precise spatial control via mask-prompt pairs. This approach offers greater flexibility and the capability to perform complex edits involving mask intersections or overlaps, surpassing the limitations of existing methods. Furthermore, it presents a computationally advantageous alternative to sequential editing techniques.
	\item To realize this editing framework, we propose CACM, which modulates both cross-attention and self-attention layers within the diffusion model. This mechanism empowers fine-grained control over the editing process, ensuring accurate adherence to user-provided instructions, without necessitating any training or test-time optimization of the base model.
	\item We contribute MiE-Bench, a new benchmark dataset specifically designed for evaluating and comparing multi-instruction image editing methods. This dataset serves as a valuable resource for advancing research in this emerging domain.
\end{itemize}

\section{Related Work}
\label{sec:relatedwork}
\textbf{Diffusion Models:} Denoising Diffusion Probabilistic Models (DDPMs) \cite{ddpm_ho_neurips2020} have demonstrated remarkable image synthesis quality, outperforming previous generative models by a substantial margin \cite{diffusion_beats_gans_neurips2021}. Consequently, various text-to-image foundation models, including DALL$\cdot$E-2 \cite{dalle2_arxiv2022} and Imagen \cite{imagen_neurips2022}, were developed. However, to address the significant training and inference complexities, the LDM approach was introduced in \cite{stablediffusion_cvpr2022}, where the diffusion model is trained within the compressed latent space of a VAE. The SD family of models \cite{stablediffusion_cvpr2022, sdxl_arxiv2023} is based on LDM.

\textbf{IBE:} In terms of text-prompt guided image editing, InstructPix2Pix (IP2P) \cite{ip2p_cvpr2023} was a pioneering work. The authors utilized Prompt-to-Prompt \cite{p2p_iclr2023} and DDIM inversion \cite{ddim_inversion_iclr2021} to generate source and edited image pairs by replacing the cross-attention maps of common tokens in the edited image reverse process. IP2P \cite{ip2p_cvpr2023} generated approximately $450,000$ image and caption pairs, capable of performing a wide range of simple and complex edits. MagicBrush \cite{magicbrush_neurips2023} further improved IP2P's performance by manually curating $10,000$ image caption pairs for fine-tuning. Other approaches such as GLIDE \cite{glide_icml2022} and SmartBrush \cite{smartbrush_cvpr2023} trained Stable Diffusion \cite{stablediffusion_cvpr2022} for mask and text-guided image editing. Blended Latent Diffusion \cite{blended_latent_diffusion_siggraph2023} attempted zero-shot image editing using a mask and an associated text instruction, but it is unreliable and requires test-time optimization similar to \cite{blended_diffusion_cvpr2022} for more reliable edits.

\section{Proposed Method}
\label{sec:proposedmethod}

\begin{figure*}[th!]
	\centering
	\includegraphics[width=0.99\textwidth]{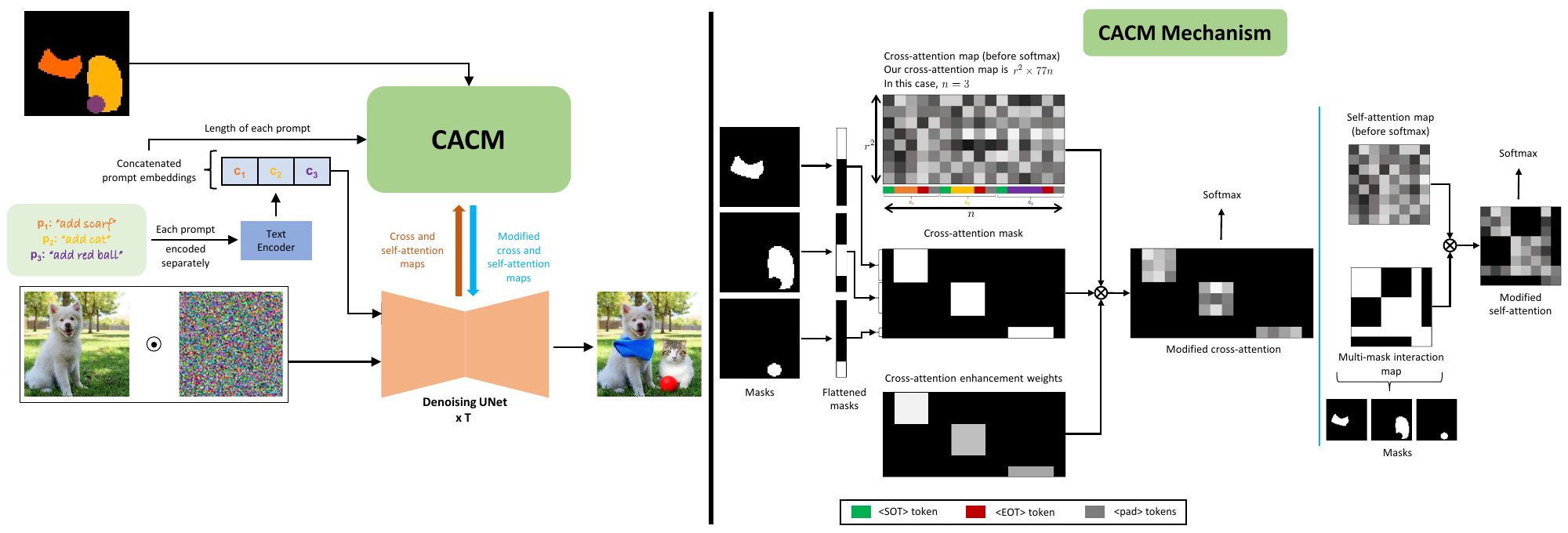}
	\caption{Overall framework of the proposed method PromptArtisan. CACM Mechanism is the backbone of PromptArtisan, enabling users to achieve fine-grained control and achieve complex edits.}
	\label{fig:proposedmethod}
	\vspace{-10pt}
\end{figure*}

PromptArtisan requires no training or test-time optimization, relying solely on the base model's capabilities. We utilize InstructPix2Pix (IP2P) \cite{ip2p_cvpr2023}. Given an image and mask-prompt pairs, PromptArtisan edits the image in one pass, enabling flexible control through intersecting or overlapping masks (Fig.~\ref{fig:teaser}). Our approach (see Fig.~\ref{fig:proposedmethod}, Alg.~\ref{algo:promptartisan}) includes managing multiple mask-prompt pairs, implementing CACM to control cross-attention and self-attention layers, and context blending for seamless integration of modified regions.

\setlength{\textfloatsep}{0.1cm}
\setlength{\floatsep}{0.1cm}
\begin{algorithm}
	\caption{PromptArtisan}
	\label{algo:promptartisan}
	\begin{algorithmic}[1]
		\renewcommand{\algorithmicrequire}{\textbf{Input:}}
		\renewcommand{\algorithmicensure}{\textbf{Output:}}
		\REQUIRE Input image: $\mathcal{I}$; Masks: $\mathcal{M} = \{m_1, m_2, ..., m_n\}$; Prompts: $\mathcal{P} = \{p_1, p_2, ..., p_n\}$; Mask order: $\mathcal{M_{\text{o}}} = \{o_1, o_2, ..., o_n\}$; Pre-trained diffusion model: $\mathcal{\epsilon_\theta(\cdot, \cdot, \cdot)}$; VAE: $\mathcal{E(\cdot)}, \mathcal{D(\cdot)}$; Text encoder: $\mathcal{T}$; Diffusion time-steps: $T$; Forward sampling process $q(z_t|z)$; Step $S$, latent blending is performed till step $S$ 
		\ENSURE  Edited image $\mathcal{I_{\text{edit}}}$
		%\\ \textit{Initialisation} :
		\STATE $z_{T} \sim \mathcal{N}(0, I) $, $z \leftarrow \mathcal{E(\mathcal{I})}$, $c_1 = c_2 = .. = c_n = \phi$
		%\STATE 
		%\STATE  
		\FOR {$i = 1$ to $n$}
		\STATE $c_i \leftarrow \mathcal{T}(p_i)$
		\ENDFOR
		\STATE $c, c_{sep} \leftarrow \text{ConcatPrompts}(c_1, c_2, ..., c_n)$
		\STATE $\mathcal{M_{\text{comp}}} = \text{CompositeMask}(\mathcal{M}, \mathcal{M_{\text{o}}})$
		\FOR {$t = T$ to $1$}
		\STATE $z_{t-1} \leftarrow \text{CACM}(\epsilon_\theta(z_t, t, z, c), \mathcal{M_{\text{comp}}}, c, c_{sep})$
		\IF {($t > S$)}
		\STATE $z_{t-1} = z_{t-1} * \mathcal{M_{\text{comp}}} + q(z_{t-1}|z) * (1-\mathcal{M_{\text{comp}}})$
		\ENDIF
		\ENDFOR
		\STATE $\mathcal{I_{\text{edit}}} = \mathcal{D}(z_0)$
		\RETURN $\mathcal{I_{\text{edit}}}$ 
	\end{algorithmic}
\end{algorithm}

\subsection{Multi-prompt Embedding Generation}
\label{subsec:multipromptembeddings}
PromptArtisan operates on mask-prompt pairs as inputs, where masks are denoted as $\mathcal{M} = \{m_1, m_2, ..., m_n\}$; Prompts: $\mathcal{P} = \{p_1, p_2, ..., p_n\}$. It is crucial to note that the embeddings for each prompt must be calculated individually. This ensures that the embedding of a given prompt remains isolated from any contextual information related to other prompts, preserving its intended meaning. We calculate the prompt embeddings separately before combining them into a single sequence of length $77n$ (see Alg.~\ref{algo:promptartisan}), reflecting the fixed token length of $77$ imposed by the CLIP encoder \cite{clip_icml2021}.

\subsection{Complete Attention Control Mechanism}
\label{subsec:crossattentioncontrol}
The CACM mechanism has two main motivations. \textbf{First}, in case of multiple instructions, it is important to remove the impact of an instruction on irrelevant regions. A given prompt should concentrate only inside its corresponding mask region. \textbf{Second}, it is important to control the self-attention between different masked regions to reduce the interference of different contents that are being generated. Our ablation study (see Fig.~\ref{fig:ablationqualitativeresults}) reveals that self-attention control plays an important role in successful prompt adherence and content generation.

\subsubsection{Cross-attention Control}
\label{subsubsec:crossattentioncontrol}
Cross-attention control, as illustrated in Fig.~\ref{fig:proposedmethod}, restricts the interaction of tokens to their respective masks. Furthermore, similar to \cite{ediff_arxiv2022}, we enhance the cross-attention scores of tokens for a given mask-prompt pair to facilitate reliable content generation. Notably, we exclude the enhancement of attention scores for the \verb|[SOT]| and \verb|[PAD]| tokens, as the start token is known to promote background generation \cite{layout_control_wacv2024}.

\subsubsection{Self-attention Control}
\label{subsubsec:selfattentioncontrol}
Self-attention control, as depicted in Fig.~\ref{fig:proposedmethod}, inhibits interactions between different masks to ensure robust and desired content generation. Our ablation study (illustrated in Fig.~\ref{fig:ablationqualitativeresults}) demonstrates that allowing a mask's pixels to attend to pixels in another mask adversely affects the content generation process.

\section{Results and Discussion}
\label{sec:resultsanddiscussion}

\subsection{Experimental Setup}
\label{subsec:trainingmethod}
To assess the efficacy of PromptArtisan, we compare it with contemporary text and mask-based IBE methods, namely GLIDE \cite{glide_icml2022}, Blended Latent Diffusion \cite{blended_latent_diffusion_siggraph2023}, DiffEdit \cite{diffedit_iclr2023}, and IP2P \cite{ip2p_cvpr2023}. To broaden the scope of comparison, we additionally compare with methods that use the technique of inversion \cite{ddim_inversion_iclr2021}. More specifically, we compare with Imagic \cite{imagic_cvpr2023} and Null-text Inversion (NTI) \cite{nulltextinv_editing_cvpr2023}. Utilizing our MiE-Bench dataset consisting of $30$ test samples sourced from \cite{pexels,unsplash}, we conduct a comprehensive evaluation and comparison of PromptArtisan against these state-of-the-art techniques. Through thorough qualitative, quantitative, and subjective analysis, we demonstrate the superiority and effectiveness of PromptArtisan.

\subsection{Qualitative Results}
\label{subsec:qualitativeresults}
\begin{figure*}[t!]
	\centering
	\includegraphics[width=0.90\textwidth]{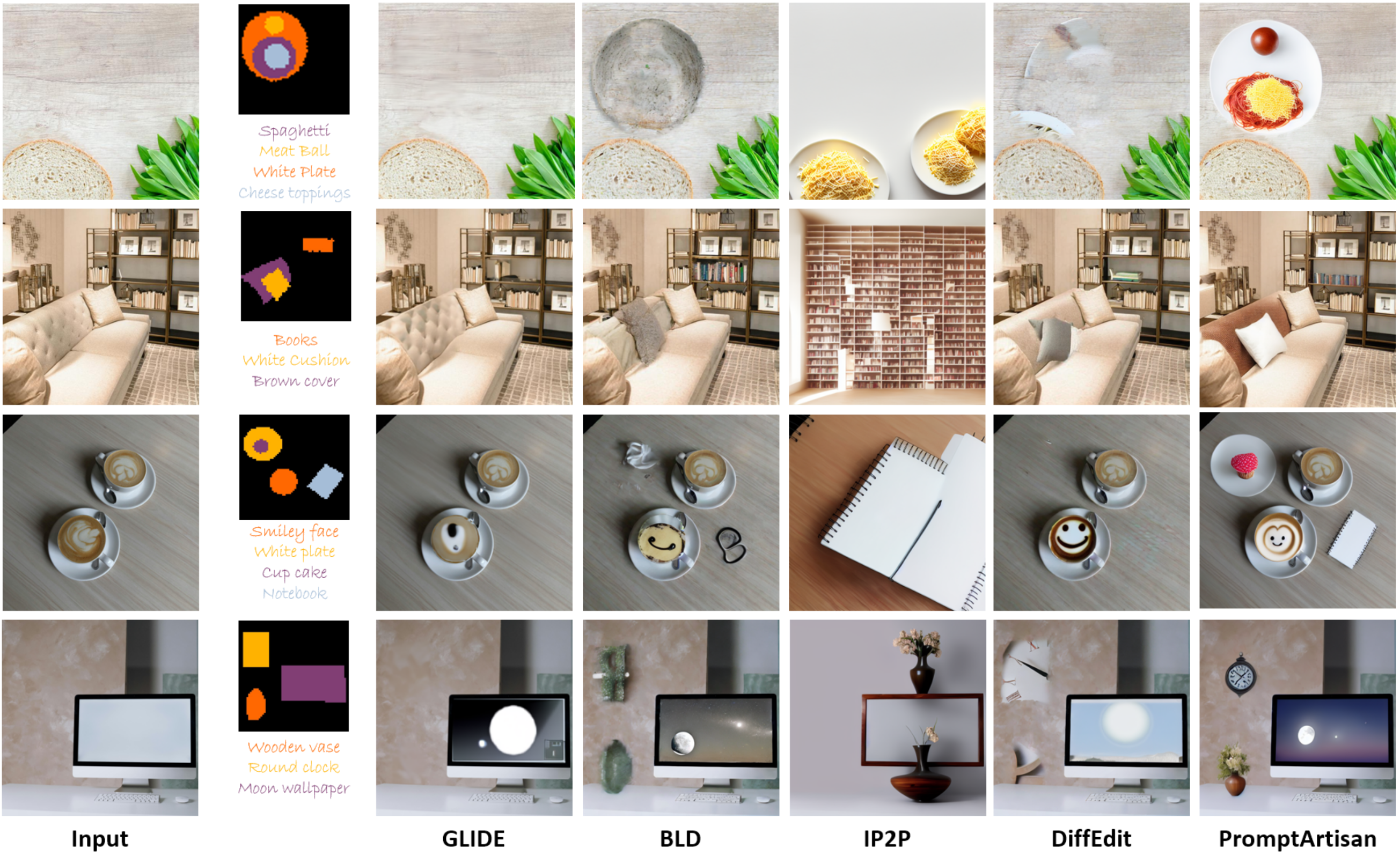}
	\caption{Qualitative comparison of PromptArtisan with competitors.}
	\label{fig:qualitativecomparison}
	\vspace{-14pt}
\end{figure*}

In our user study, $10$ participants evaluated the visual quality and instruction adherence of edited images. Table~\ref{tab:subjectivestudyresults} presents the summarized results. PromptArtisan consistently ranked top in terms of visual characteristics of the final edited image, implied by its overall preference by users. Specifically, participants appreciated PromptArtisan's ability to handle complex instructions and overlapping masks effectively. As exemplified in row $1$ and $3$ of Fig.~\ref{fig:qualitativecomparison}, PromptArtisan successfully preserved image fidelity while adhering to user instructions, even in challenging scenarios involving the addition of overlapping objects.

\subsection{Quantitative Results}
\label{subsec:quantitativeresults}
In order to provide a comprehensive assessment of PromptArtisan, we also consider quantitative metrics such as CLIP score and PickScore \cite{pickscore_neurips2023}. These metrics are commonly employed to gauge the degree to which a generated image aligns with the user's intention (prompt). In the context of IBE, higher values of these metrics signify a stronger adherence to user instructions during editing. As evident from Table~\ref{tab:quantitativecomparison}, PromptArtisan attains the highest CLIP Score and PickScore, demonstrating its proficiency in faithfully executing user instructions compared to existing methods.

\begin{table}[t!]
	\caption{Quantitative comparison of PromptArtisan with competitors.}
	\label{tab:quantitativecomparison}
	\centering
	\renewcommand{\arraystretch}{1.15}
	\setlength{\tabcolsep}{0.45em}
	\resizebox{0.50\textwidth}{!}{
		\begin{tabular}{c|c|c|c|c|c|c|c}
			\hline
			\textbf{} & \textbf{GLIDE} & \textbf{BLD}& \textbf{IP2P}& \textbf{DiffEdit} & \textbf{Imagic} & \textbf{NTI} & \textbf{PromptArtisan}			\\
			\hline
			\textbf{CLIP Score} & 	$22.08$ 	& 	$27.02$ 	& 	$27.79$ 	& 	$23.55$ 	& 	$17.02$ 	& 	$27.74$		&		$\textbf{28.78}$	\\ 
			\textbf{PickScore} 	& 	$0.0876$ 	&	$0.1371$	&  	$0.1288$ 	&	$0.1138$	& 	$0.1025$ 	& 	$0.1070$	&		$\textbf{0.3230}$	\\
			\hline
		\end{tabular}
	}
\end{table}

\begin{table}[t!]
	\caption{Results of the user study to judge the aesthetic qualities of the edited image.}
	\label{tab:subjectivestudyresults}
	\centering
	\renewcommand{\arraystretch}{1.15}
	\setlength{\tabcolsep}{0.45em}
	\resizebox{0.48\textwidth}{!}{
		\begin{tabular}{c|c|c|c|c|c|c|c}
			\hline
			\textbf{} & \textbf{GLIDE} & \textbf{BLD}& \textbf{IP2P}& \textbf{DiffEdit} & \textbf{Imagic} & \textbf{NTI} & \textbf{PromptArtisan}			\\
			\hline
			\textbf{Pref. rate} &  	$1.39$ 		& 	$2.54$ 		& 	$1.77$ 		& 	$1.65	$ 	& 	$1.22$ 		& 	$2.41$		&		$\textbf{4.35}$		\\
			\hline
		\end{tabular}
	}
\end{table}

\subsection{Ablation Study}
\label{subsec:ablationstudy}
\begin{figure}[t!]
	\centering
	\includegraphics[width=0.45\textwidth]{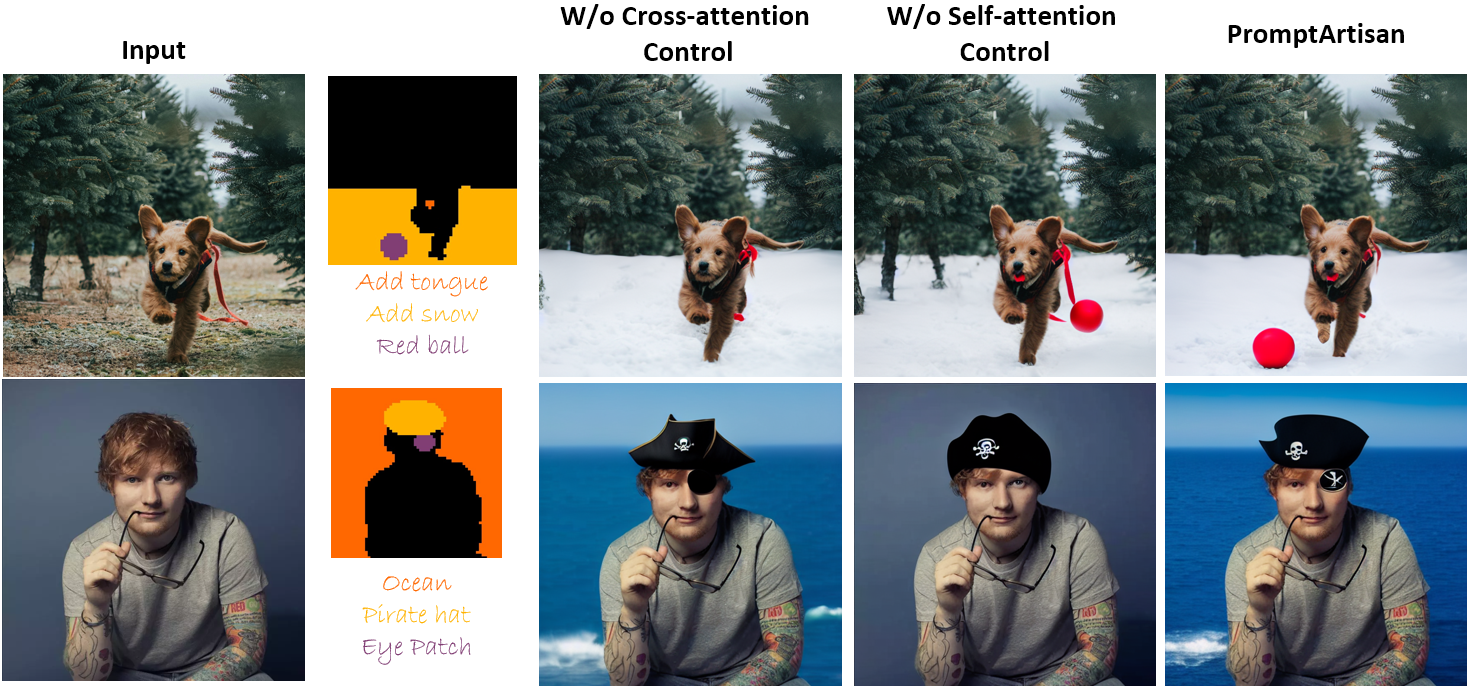}
	\caption{Qualitative results of our ablation study.}
	\label{fig:ablationqualitativeresults}
\end{figure}

To highlight the significance of self and cross attention control in PromptArtisan, we conducted an ablation study. The results of the study are depicted in Fig.~\ref{fig:ablationqualitativeresults}.

\subsection{Additional Results}
\label{subsec:additionalresults}
\begin{figure}[t!]
	\centering
	\includegraphics[width=0.42\textwidth]{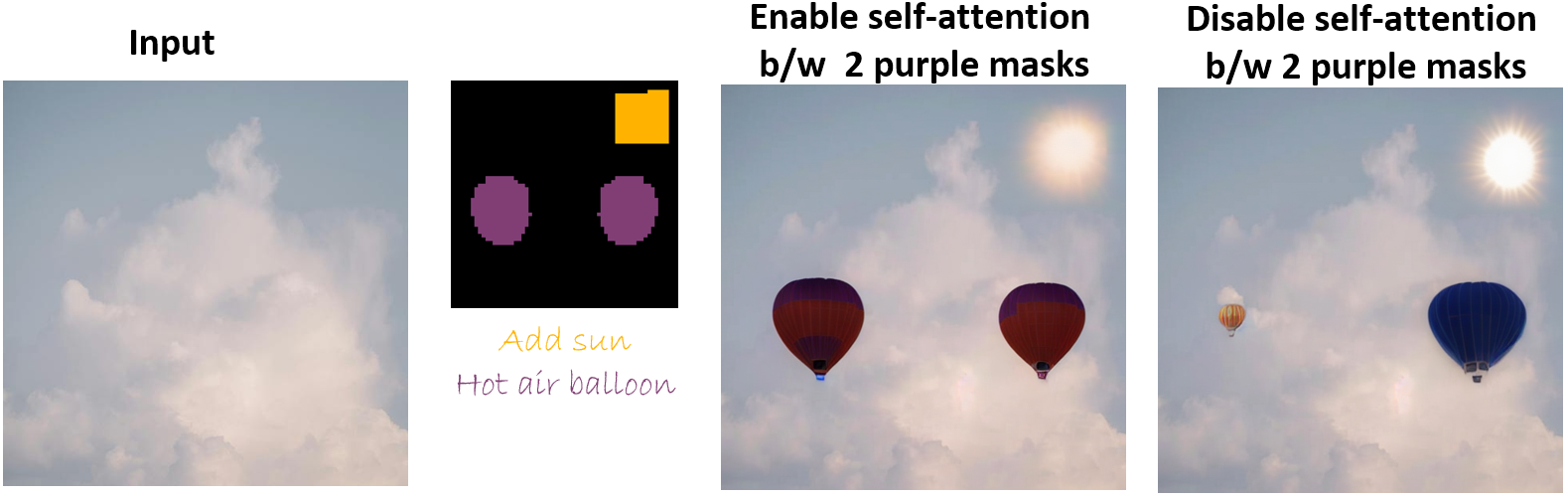}
	\caption{Additional capabilities of PromptArtisan.}
	\label{fig:additionalresults}
\end{figure}

PromptArtisan's capability to add objects in multiple areas simultaneously is noteworthy in the IBE domain. As observed in Fig.~\ref{fig:additionalresults}, PromptArtisan successfully inserts the same object when the masks share the same color (self-attention is enabled between these masks). This feature paves the way for more sophisticated editing scenarios involving repeated objects.

\section{Conclusion and Future Work}
\label{sec:conclusion}
PromptArtisan represents a significant breakthrough in multi-instruction image editing, providing unparalleled flexibility and control to users. With its innovative approach, PromptArtisan eliminates the need for time-consuming iterative refinement, empowering users to achieve remarkable results in a single pass. By leveraging a pre-trained InstructPix2Pix model and incorporating a novel complete attention control mechanism, PromptArtisan grants users fine-grained control over the editing process, ensuring precise adherence to their instructions. Moreover, PromptArtisan's zero-shot nature also sets it apart. Overall, PromptArtisan opens up exciting new possibilities for creative and efficient image editing workflows, catering to users of all skill levels.

\newpage
\bibliographystyle{IEEEtran}
\bibliography{references}

\end{document}